\documentstyle[jair,twoside,11pt,theapa]{article}
\input epsf 

\jairheading{12}{2000}{93-103}{11/99}{3/00}
\ShortHeadings{Exact Phase Transitions in Random Constraint Satisfaction 
Problems}
{Xu \& Li}
\firstpageno{93}
\newtheorem{theorem}{\bf Theorem}

\newtheorem{definition}{\bf Definition}

\newcommand{\comb}[2]{\Big(\begin{array}{c}\small #1\\ #2 \end{array}\Big)}

\begin{document}

\title{
Exact Phase Transitions in\\
Random Constraint
Satisfaction Problems
}

\author{\name Ke Xu    \email kexu@nlsde.buaa.edu.cn\\
\name Wei Li   \email liwei@nlsde.buaa.edu.cn\\
\addr National Laboratory of Software Development Environment,\\
Department of Computer Science and Engineering,\\
Beijing University of Aeronautics and Astronautics,\\
Beijing, 100083, P.R. China}

\maketitle
\begin{abstract}
In this paper we propose a new type of random CSP model, 
called Model RB, which is a revision to the standard Model B.
It is proved that phase transitions from 
a region where almost all problems 
are satisfiable to a region where almost all problems
are unsatisfiable do exist for Model RB 
as the number of variables approaches infinity.
Moreover, the critical values at which 
the phase transitions occur are also
known exactly. By relating the hardness of Model RB to Model B, 
it is shown that there exist a lot of hard instances in Model RB.
\end{abstract}

\section{Introduction}
\label{Introduction}

Since the seminal paper of Cheeseman, Kanefsky and Taylor  
\citeyear{cheeseman} appeared, 
there has been a great amount of interest in the study of phase transitions
in NP-complete problems. However, it seems to be very difficult to prove 
the existence of this phenomenon or to obtain the exact location
of the transition points for such problems. For example, in random $3$-SAT, 
it is known from experiments that the phase transition will occur 
when the ratio of clauses to variables is approximately $4.3$ 
\cite{mitchell}. Another experimental estimate of the
transition point suggested by Kirkpatrick and Selman \citeyear{kirkpatrick}
is $4.17$. 
They used finite-size scaling methods from statistical physics to
derive the result. In contrast with the experimental studies, the
theoretical work has only given some loose but hard won bounds on the
location of the transition point. Currently, the best known lower bound 
and upper bound are $3.003$ \cite{frieze}
and $4.602$ \shortcite{kirousis} respectively. 
Recently, Friedgut \citeyear{friedgut} made tremendous progress 
towards establishing 
the existence of a threshold for random $k$-SAT by proving that the
width of the transition region narrows as the number of variables increases.
But we still can not obtain the exact location of the phase transition point 
from this approach.

In fact, SAT is a special case of the constraint satisfaction problem (CSP). 
CSP has not only important theoretical value in artificial intelligence, 
but also many immediate applications in areas ranging from vision, 
language comprehension to scheduling and diagnosis \cite{dechter}. In
general, CSP tasks are computationally intractable (NP-hard) \cite{dechter}. 
In recent years random constraint satisfaction problems have also received 
great attention, both from an experimental and a theoretical point of view 
\shortcite{achlioptas,cheeseman,frost,gent,hogg,larrosa,prosser,purdom,smith,smitha,williams}.
Williams and Hogg \citeyear{williams} 
developed a technique to predict where the hardest 
problems are to be found and where the fluctuations in difficulty are greatest
in a space of problem instances. They have also shown that
their predictions of the critical value agree well with the experimental data. 
Smith and Dyer \citeyear{smith} studied the location of 
the phase transition in binary constraint satisfaction problems and discussed
the accuracy of a prediction based on the expected number of solutions. 
Their results show that the variance of the number of solutions can be 
used to set bounds on the phase transition and to
indicate the accuracy of the prediction. 
Recently, a theoretical result by Achlioptas et al. \citeyear{achlioptas}
shows that many models commonly used for 
generating random CSP instances do not
have an asymptotic threshold due to the presence of flawed variables. 
More recently, Gent et al. \citeyear{gent}
have shown how to introduce structure into the conflict matrix to 
eliminate flaws.

In this paper we propose a new type of random CSP model, 
called Model RB, which is a revision to the standard Model B 
\cite{gent,smith}. It is proved that the
phase transition phenomenon does exist for Model RB as the number of 
variables approaches infinity. 
More precisely, there exist two control parameters $r$, $p$ 
and the corresponding critical values $r_{cr}$, $p_{cr}$    
such that for each fixed value $r<r_{cr}$ or $p<p_{cr}$,
a random CSP instance
generated following Model RB is satisfiable with probability tending to $1$ 
as the number of variables approaches infinity, and when $r>r_{cr}$
or $p>p_{cr}$, unsatisfiable with probability tending to $1$. 
Moreover, the critical values $r_{cr}$  and $p_{cr}$  are also known exactly. 
By relating the hardness of Model RB to Model B, 
it is shown that Model RB actually has a lot of hard instances.

\section{Definitions and Notations}
\label{definitions}

A {\em constraint satisfaction problem} (CSP) consists of a finite set
$U=\{u_1,\cdots,u_n\}$ of $n$  variables and a set of constraints.
For each variable $u_i$, a {\em domain} $D_i$ with $d_i$  
elements is specified; a variable can only be assigned a value from its 
domain. For $2\leq k\leq n$  a {\em constraint} $C_{i1,i2,\cdots,ik}$
consists of a subset $\{u_{i1},u_{i2},\cdots,u_{ik}\}$  of $U$ 
and a relation 
$R_{i1,i2,\cdots,ik}\subseteq D_{i1}\times\cdots\times D_{ik}$, 
where $i1,i2,\cdots,ik$ are distinct. 
$C_{i1,i2,\cdots,ik}$  is called a $k$-ary constraint which 
bounds the variables $u_{i1},\cdots,u_{ik}$. $R_{i1,i2,\cdots,ik}$
specifies all the allowed tuples of values for the variables 
$u_{i1},\cdots,u_{ik}$  which are compatible with each other. 
A {\em solution} to a CSP is an assignment of a value
to each variable from its domain such that all the constraints 
are satisfied. A constraint $C_{i1,i2,\cdots,ik}$  is satisfied if the
tuple of values assigned to the variables $u_{i1},\cdots,u_{ik}$  
is in the relation $R_{i1,i2,\cdots,ik}$ . A CSP that has  a
solution is called {\em satisfiable}; otherwise it is {\em unsatisfiable}.
In this paper, the probability of a random CSP instance being satisfiable
is denoted by $Pr(Sat)$.

We assume that $k\geq 2$ and all the variable domains contain 
the same number of values $d=n^\alpha$ in Model RB (where $\alpha$ 
is a constant). The generation of a random CSP instance in
Model RB is done in the following two steps:\\
{\em 
Step 1.  We select with repetition $t=rn\ln n$  random constraints. 
Each random constraint is formed by selecting without repetition 
$k$  of $n$  variables.\\
Step 2.  For each constraint we uniformly select without repetition 
$q=p\cdot d^k$  incompatible tuples of values, i.e., each constraint 
relation contains exactly $(1-p)\cdot d^k$ compatible tuples of values.
}

The parameter $r$  determines how many constraints are in a CSP instance, 
while $p$ determines how restrictive the constraints are.

The following definitions will be needed in section 4 when we derive 
the expectation of the second moment of the number of solutions.
\begin{definition}
An assignment pair is an ordered pair $\langle t_i,t_j\rangle$  
of assignments to the variables in $U$, where
$t_i=(a_{i1},a_{i2},\cdots,a_{in})$ and $t_j=(a_{j1},a_{j2},\cdots,a_{jn})$
with $a_{il},a_{jl}\in D_l$. An assignment pair $\langle t_i,t_j\rangle$
satisfies a CSP if and only if both $t_i$  and $t_j$  satisfy this CSP.
The set that consists of all the assignment pairs is denoted by $A_{pair}$.
\end{definition}
\begin{definition}
Similarity number $S^f:A_{pair}\mapsto\{0,1,2,\cdots\}$,
\begin{equation}\small
S^f(\langle t_i,t_j\rangle)=\sum_{l=1}^n Sam(a_{il},a_{jl})
\end{equation} 
where the function $Sam$  is defined as follows:
\begin{equation}\small
Sam(a_{il},a_{jl})=\left\{
\begin{array}{ll}\small
1&\mbox{ if } a_{il}=a_{jl}\\
0&\mbox{ if } a_{il}\not= a_{jl}
\end{array}
\right.
\end{equation}
\end{definition}

The similarity number of an assignment pair is equal to the number of 
variables at which the two assignments of this assignment pair take the 
identical values. By Definition 2 it is easy to see
that $0\leq S^f(\langle t_i,t_j\rangle)\leq n$.

\section{Main Results}
\label{main}

In this paper, the following theorems are proved.
\begin{theorem}
Let $r_{cr}=-\frac{\alpha}{\ln(1-p)}$. 
If $\alpha>\frac{1}{k}$, $0<p<1$ are two constants and
$k$, $p$ satisfy the inequality $k\geq \frac{1}{1-p}$, then
\begin{equation}\small
\lim_{n\rightarrow\infty}Pr(Sat)=1 \mbox{ when } r<r_{cr}
\end{equation}
\begin{equation}\small
\lim_{n\rightarrow\infty}Pr(Sat)=0 \mbox{ when } r>r_{cr}
\end{equation}
\end{theorem}
\begin{theorem}
Let $p_{cr}=1-e^{-\frac{\alpha}{r}}$. 
If $\alpha>\frac{1}{k}$, $r>0$  are two constants and
$k$, $\alpha$ and $r$  satisfy the inequality  
$ke^{-\frac{\alpha}{r}}\geq 1$, then
\begin{equation}\small
\lim_{n\rightarrow\infty}Pr(Sat)=1 \mbox{ when } p<p_{cr}
\end{equation}
\begin{equation}\small
\lim_{n\rightarrow\infty}Pr(Sat)=0 \mbox{ when } p>p_{cr}
\end{equation}
\end{theorem}

\section{Proof of Theorem 1 and Theorem 2}
\label{proof}

The expected number of solutions $E(N)$ for model RB is given by
\begin{equation}\small
E(N)=d^n (1-p)^{rn\ln n}=n^{\alpha n}(1-p)^{rn\ln n}
\end{equation}
By the Markov inequality $Pr(Sat)\leq E(N)$  it is not hard to 
show that $\lim_{n\rightarrow\infty}Pr(Sat)=0$   when
$r>r_{cr}$  or $p>p_{cr}$. Hence relations (4), (6) are proved.
It is also easy to see that $E(N)$ is eqal to 1 
when $r=r_{cr}$  or $p=p_{cr}$, and $E(N)$  grows exponentially
with $n$  when $r<r_{cr}$  or $p<p_{cr}$.

The key point in the proof of relations (3), (5) 
is to derive the expectation of the second
moment $E(N^2)$  and give an asymptotic estimate of it. 
Let $\phi$ be a random CSP instance generated following Model RB. 
$P(\langle t_i,t_j\rangle)$  stands for the probability of 
$\langle t_i,t_j\rangle$  satisfying $\phi$.
Now we start to derive the expression of $P(\langle t_i,t_j\rangle)$. 
Since each constraint is generated
independently, we only need to consider the probability of 
$\langle t_i,t_j\rangle$  satisfying a random constraint. 
Assuming that the similarity number of $\langle t_i,t_j\rangle$
is equal to $S$, we have the following two cases:

(1) Each variable of a constraint is assigned the same value in
$t_i$ as that in $t_j$.  In this case,
the probability of $\langle t_i,t_j\rangle$  satisfying the constraint is
$\comb{d^k-1}{q}/\comb{d^k}{q}$.

(2) Otherwise, the probability of $\langle t_i,t_j\rangle$ 
 satisfying a constraint is  $\comb{d^k-2}{q}/\comb{d^k}{q}$.

The probability that a random constraint falls into the first case is
$\comb{S}{k}/\comb{n}{k}$.
Hence the probability into the second case is
$1- \comb{S}{k}/\comb{n}{k}$.
Thus we get
\begin{equation}\small
P(\langle t_i,t_j\rangle)=
\left(
\frac{\comb{d^k-1}{q}}{\comb{d^k}{q}}
\cdot
\frac{\comb{S}{k}}{\comb{n}{k}}
+
\frac{\comb{d^k-2}{q}}{\comb{d^k}{q}}
\cdot 
(
1-\frac{\comb{S}{k}}{\comb{n}{k}}
)
\right)^{rn\ln n}
\end{equation}

Let $A_S$  be the set of assignment pairs whose {\em similarity number}
is equal to $S$ . It is easy to show that the cardinality of $A_S$
is given by
\begin{equation}\small
|A_S|=d^n\comb{n}{S} (d-1)^{n-S}
\end{equation} 

>From the definition of $E(N^2)$, we have
\begin{eqnarray}\small
E(N^2)&=&\sum_{S=0}^n |A_S|P(t_i,t_j)\nonumber\\
&=&d^n\comb{n}{S} (d-1)^{n-S}
\left(
\frac{\comb{d^k-1}{q}}{\comb{d^k}{q}}\cdot
\frac{\comb{S}{k}}{\comb{n}{k}}+
\frac{\comb{d^k-2}{q}}{\comb{d^k}{q}}\cdot
(
1-\frac{\comb{S}{k}}{\comb{n}{k}}
)
\right)^{rn\ln n}
\end{eqnarray}

It is very difficult to analyze the above expression directly. 
First, we give an asymptotic estimate of $P(\langle t_i,t_j\rangle)$. 
Let $s=\frac{S}{n}$. It is obvious that $0\leq s\leq 1$. 
By asymptotic analysis, we get
$$
\frac{\comb{S}{k}}{\comb{n}{k}}= \frac{
\frac{S}{n}(\frac{S}{n}-\frac{1}{n})
(\frac{S}{n}-\frac{2}{n})\cdots
(\frac{S}{n}-\frac{k-1}{n})
}{
(1-\frac{1}{n})(1-\frac{2}{n})\cdots
(1-\frac{k-1}{n})
}=
s^k+\frac{g(s)}{n}+O(\frac{1}{n^2})
$$ 
where
\begin{equation}\small
g(s)=\frac{k(k-1)(s^k-s^{k-1})}{2}
\end{equation}
and
\begin{equation}\small
\frac{\comb{d^k-1}{q}}{\comb{d^k}{q}}=
\frac{d^k-q}{d^k}=1-p
\end{equation}
\begin{equation}\small
\frac{\comb{d^k-2}{q}}{\comb{d^k}{q}}=
\frac{(d^k-q)(d^k-q-1)}{d^k(d^k-1)}
=(1-p)^2+O(\frac{1}{d^k})
\end{equation}
Note that $d=n^\alpha$, we have
\begin{equation}\small
P(\langle t_i,t_j\rangle)=
\left[
(1-p)\cdot(s^k+\frac{g(s)}{n})+
(1-p)^2\cdot(1-s^k-\frac{g(s)}{n})+
O(\frac{1}{n^2})+O(\frac{1}{n^{k\alpha}})
\right]^{rn\ln n}
\end{equation}
By use of the condition $\alpha>\frac{1}{k}$, we get
\begin{equation}\small
P(\langle t_i,t_j\rangle)=
(1-p)^{2rn\ln n}\left[
1+\frac{p}{1-p}(s^k+\frac{g(s)}{n})
\right]^{rn\ln n}(1+O(\frac{1}{n}))
\end{equation} 

For every $0<s<1$ (where $s=\frac{S}{n}$), 
the asymptotic estimate of $|A_S|$  is
\begin{eqnarray}\small
|A_S|&=&n^{\alpha n}(n^\alpha-1)^{n-ns}
\frac{1}{\sqrt{2\pi n s(1-s)}}
e^{n(-s\ln s -(1-s)\ln (1-s))}(1+O(\frac{1}{n}))\nonumber\\
&=&n^{2\alpha n}(1-\frac{1}{n^{\alpha}})^{n-ns}
(\frac{1}{n^{\alpha}})^{ns}
\frac{1}{\sqrt{2\pi n s(1-s)}}
e^{n(-s\ln s -(1-s)\ln(1-s))}(1+O(\frac{1}{n}))
\end{eqnarray} 
Notice that $E(N)=n^{\alpha n}(1-p)^{rn\ln n}$, we have
\begin{equation}\small
|A_S|P(\langle t_i,t_j\rangle)=E^2(N)\left[
1+\frac{p}{1-p}(s^k+\frac{g(s)}{n})
\right]^{rn\ln n}
(1-\frac{1}{n^\alpha})^{n-ns}(\frac{1}{n^\alpha})^{ns}
\comb{n}{ns} (1+O(\frac{1}{n}))
\end{equation}

When $n$ is sufficiently large, except the first term $E^2(N)$, 
$|A_S|P(\langle t_i,t_j\rangle)$ is mainly
determined by the following terms:
\begin{equation}
f(s)=\left[
1+\frac{p}{1-p}s^k
\right]^{rn\ln n}
(\frac{1}{n^\alpha})^{ns}
\end{equation}
We can rewrite it as
\begin{equation}
f(s)=e^{
\left[
r\ln(1+\frac{p}{1-p}s^k)-\alpha s
\right]n\ln n
}
\end{equation}         
Let
\begin{equation}
h(s)=r\ln(1+\frac{p}{1-p}s^k)-\alpha s
\end{equation} 
The second derivative of $h(s)$  is
\begin{equation}
h''(s)=\frac{
rkps^{k-2}[(k-1)(1-p)-ps^k]
}{
(1-p+ps^k)^2
}
\end{equation} 
Applying the condition $k\geq\frac{1}{1-p}$
in Theorem 1 to the above equation we can easily prove
that $h''(s)\geq 0$ on the interval $0\leq s\leq 1$.
For Theorem 2, from the condition $ke^{-\frac{\alpha}{r}}\geq 1$
it follows that the inequality $k\geq\frac{1}{1-p}$
still holds when $p<p_{cr}$. It is also not hard to show that 
$h(0)=0$,
and $h(1)=-r\ln(1-p)-\alpha<0$  when $r<r_{cr}$  or $p<p_{cr}$. 
Hence we can easily prove that the
unique maximum point of $h(s)$  is $s=0$ 
when $r<r_{cr}$  or $p<p_{cr}$ . Thus the terms of $0<s\leq 1$
are negligible when $r<r_{cr}$  or $p<p_{cr}$ . 
We only need to consider those terms near $s=0$ . The
process can be divided into the following three cases:

{\bf Case 1:} $\alpha>1$.
When $S=0$ ($s=0$), from the definition of $g(s)$  in Equation (11) we have
\begin{equation}
\left[
1+\frac{p}{1-p}(s^k+\frac{g(s)}{n})
\right]^{rn\ln n}=1
\end{equation}
Thus by Equation (17) we get
\begin{equation}
|A_S|P(\langle t_i,t_j\rangle)\approx
E^2(N)(1-\frac{1}{n^\alpha})^n\approx
E^2(N)
\end{equation} 
When $S=1$ ($s=\frac{1}{n}$), it also not hard to prove that
\begin{equation}
\lim_{n\rightarrow\infty}\left[
1+\frac{p}{1-p}(s^k+\frac{g(s)}{n})
\right]^{rn\ln n}=e^0=1
\end{equation} 
Hence we obtain
\begin{equation}
|A_S|P(\langle t_i,t_j\rangle)\approx
E^2(N)n^{1-\alpha}\mbox{ when } S=1 
\end{equation}
Similary,
$$
|A_S|P(\langle t_i,t_j\rangle)\approx
E^2(N)\frac{n^{2(1-\alpha)}}{2!}\mbox{ when } S=2
$$
\begin{equation}
|A_S|P(\langle t_i,t_j\rangle)\approx
E^2(N)\frac{n^{3(1-\alpha)}}{3!} \mbox{ when } S=3,\cdots 
\end{equation}
Summing the above terms together, we obtain
\begin{equation}
E(N^2)=\sum_{S=0}^n |A_S|P(\langle t_i,t_j\rangle)\approx
E^2 (N)e^{n^{1-\alpha}}\approx E^2 (N)
\end{equation}
 
{\bf Case 2:} $\alpha=1$.
By use of the method in Case 1, it can be easily shown that
$$
|A_S|P(\langle t_i,t_j\rangle)\approx
E^2(N)(1-\frac{1}{n})^n\approx E^2(N)\frac{1}{e}
\mbox{ when } S=0 
$$
$$
|A_S|P(\langle t_i,t_j\rangle)\approx
E^2(N)\frac{1}{e}\mbox{ when } S=1
$$
$$
|A_S|P(\langle t_i,t_j\rangle)\approx
E^2(N)\frac{1}{e\cdot 2!}\mbox{ when } S=2
$$
\begin{equation}
|A_S|P(\langle t_i,t_j\rangle)\approx
E^2(N)\frac{1}{e\cdot 3!}\mbox{ when } S=3,\cdots
\end{equation}
Summing the above terms together, we obtain
\begin{equation}
E(N^2)=
\sum_{S=0}^n |A_S|P(\langle t_i,t_j\rangle)\approx
E^2 (N)\frac{1}{e}\cdot e=E^2 (N)
\end{equation} 

{\bf Case 3:} $\frac{1}{k}<\alpha<1$.
Let $S_0=n^\beta$ (where $\beta$  is a constant and satisfies
$1-\alpha<\beta<1-\frac{1}{k}$). It is not hard to show that
when $0\leq S\leq S_0$ ($0\leq s\leq n^{\beta-1}<n^{-\frac{1}{k}}$),
the following limit holds:
\begin{equation}
\lim_{n\rightarrow\infty}\frac{p}{1-p}(s^k+\frac{g(s)}{n})
\cdot n\ln n=0
\end{equation}
Thus when $0\leq S\leq S_0$, 
the asymptotic estimate of the second term in the right 
of Equation (17) is
\begin{equation}
\left[
1+\frac{p}{1-p}(s^k+\frac{g(s)}{n})
\right]^{rn\ln n}\approx e^0=1\mbox{ when }
n\rightarrow\infty
\end{equation}
So when $0\leq S\leq S_0$, 
the asymptotic estimate of $|A_S|P(\langle t_i,t_j\rangle)$ is
\begin{equation}
|A_S|P(\langle t_i,t_j\rangle)\approx
E^2(N)\comb{n}{S}(1-\frac{1}{n^\alpha})^{n-S}
(\frac{1}{n^\alpha})^S
\end{equation} 
It should be noted that 
$\comb{n}{S}(1-\frac{1}{n^\alpha})^{n-S}(\frac{1}{n^\alpha})^S$ 
 is a binomial term whose maximum point is around
$S=n^{1-\alpha}$, and $S_0=n^\beta>n^{1-\alpha}$.
By asymptotic analysis, we obtain
\begin{equation}
\sum_{S=0}^{S_0}
\comb{n}{S}(1-\frac{1}{n^\alpha})^{n-S}(\frac{1}{n^\alpha})^S
\approx
\sum_{S=0}^{n}
\comb{n}{S}(1-\frac{1}{n^\alpha})^{n-S}(\frac{1}{n^\alpha})^S
=1
\end{equation}
Thus we get
\begin{equation}
E(N^2)=\sum_{S=0}^n|A_S|P(\langle t_i,t_j\rangle)
\approx E^2(N)
\end{equation} 
Combining the above three cases gives
\begin{equation}
E(N^2)\approx E^2(N)\mbox{ when } r<r_{cr} \mbox{ or when } p<p_{cr}
\end{equation}
Hence
\begin{equation}
\lim_{n\rightarrow\infty}\frac{E^2(N)}{E(N^2)}=1
\mbox{ when } r<r_{cr} \mbox{ or when } p<p_{cr}
\end{equation}

By the Cauchy inequality $Pr(Sat)\geq\frac{E^2(N)}{E(N^2)}$ 
\cite{bollo}, 
it can be easily proved that
$\lim_{n\rightarrow\infty}Pr(Sat)=1$ when $r<r_{cr}$ or $p<p_{cr}$. 
Hence Theorem 1 and Theorem 2 are proved.

\section{The Relation between Model B and Model RB}
\label{relation}
In this section we will explain in detail how Model RB 
revises Model B and show the hardness of
Model RB by relating it to Model B. 
>From the previous papers \cite{gent,smith}
we know that the generation of a random CSP instance 
in the standard Model B (which is
often written as $\langle n,d,p_1,p_2\rangle$) 
is done in the following two steps:\\
{\em
Step 1.  We select with repetition $t=p_1\frac{n(n-1)}{2}$
random constraints. Each random constraint is
formed by selecting without repetition $2$ of $n$  variables.\\
Step 2.  For each constraint we uniformly select without repetition
$q=p_2\cdot d^2$  incompatible tuples of values, i.e.,
each constraint relation contains exactly 
$(1-p_2)\cdot d^2$ compatible tuples of values.
}

Since the standard Model B is a binary CSP model,
we will only consider the binary case of
Model RB in this section. 
In the previous papers Model B was used to test the CSP algorithms in
the following way. Given the values of $n$, $d$  and $p_1$, 
the constraint tightness $p_2$  was varied from $0$ to $1$
in steps of $\frac{1}{d^2}$.
At each setting of $\langle n,d,p_1,p_2\rangle$
a fixed number of instances (e.g. $100$) were generated.
The search algorithm was then applied to each instance. 
Finally numerous statistics about the search cost and 
the probability of being satisfiable were gathered. In
fact, the two steps of forming a constraint and selecting 
incompatible tuples of values in Model RB is exactly the
same as those in Model B. The significant difference between Model B and
Model RB is that Model RB restricts how fast the domain size and 
the number of constraints increase with the number of
variables while Model B does not, which may lead to the result that
many instances of Model B suffer from being asymptotically 
trivially insoluble \cite{achlioptas} while Model RB 
avoids this problem. But it is easy to see that given the values of 
$n$, $d$ and $p_1$,
for the setting $\langle n,d,p_1,p_2\rangle$ of Model B 
there is an equivalent setting in Model RB
with the same number of variables as that in $\langle n,d,p_1,p_2\rangle$,
$\alpha=\frac{\ln d}{\ln n}$ and 
$r=\frac{p_1(n-1)}{2\ln n}$ (Let
$n^\alpha=d$ and $rn\ln n=\frac{1}{2}p_1 n(n-1)$).

Theorem 2 shows that if $\alpha>\frac{1}{2}$  and  
$2e^{-\frac{\alpha}{r}}\geq 1$, then there exists an exact phase 
transition in the binary case of Model RB.
Given the values of $n$, $d$  and $p_1$  in Model B, for the setting of
$\langle n,d,p_1,p_2\rangle$, 
the conditions that the equivalent setting in Model RB satisfies Theorem 2 are
\begin{equation}
\alpha=\frac{\ln d}{\ln n}>\frac{1}{2}\Rightarrow d^2>n
\end{equation}      
\begin{equation}
2e^{-\frac{\alpha}{r}}\geq 1\Rightarrow
2e^{-\frac{\ln d}{\ln n}\cdot\frac{2\ln n}{p_1(n-1)}}
\geq 1\Rightarrow p_1\geq\frac{2\ln d}{(n-1)\ln 2}
\end{equation} 

The proof of Theorem 2 reveals that if the conditions (37), (38) 
are satisfied, then Model RB will exhibit an exact phase transition at
$E(N)=1$. It should be noted that Williams and Hogg \cite{williams},
and independently Smith \citeyear{smith} have already developed a theory
to predict the phase transition point in Model B on the basis of 
$E(N)=1$. Prosser \citeyear{prosser} found that this theory is in
close agreement with the empirical results, except when $p_1$ is small. 
Inequality (38) shows that in order to make the equivalent setting 
in Model RB satisfy the conditions of Theorem 2, the
parameter $p_1$  in Model B should not be less than a certain value, 
which is consistent with Prosser's experimental finding.

Now we consider a typical setting $\langle 20, 10, 0.5,p_2\rangle$ 
of Model B. Let $n=20$, $\alpha=\frac{\ln 10}{\ln 20}\approx 0.7686$
and $r=\frac{0.5(20-1)}{2\ln 20}\approx 1.5856$ in Model RB. 
Then the setting of Model RB with such values is equivalent to the 
setting $\langle 20, 10, 0.5,p_2\rangle$ of Model B.
>From Inequalities (37) and (38) it is also not hard to show that 
the equivalent setting in Model RB corresponding to the
setting $\langle 20, 10, 0.5,p_2\rangle$
satisfies the conditions of Theorem 2, i.e., $10^2>20$  and
$p_1=0.5\geq\frac{2\ln 10}{(20-1)\ln 2}\approx 0.35$. 
The experiments done by Prosser \citeyear{prosser} show that the instances
generated at $p_2=0.38$ are very hard to solve. 
This maximum cost point also agrees well
with the asymptotic phase transition point of Model RB that is 
$p=1-e^{-\frac{\alpha}{r}}\approx 1-e^{-\frac{0.7686}{1.5856}}\approx 0.38$. 
For some other settings of Model B in the previous work, 
we can also find their equivalent settings in Model RB using this method. 
Thus the hardness of solving these settings of Model B is
equal to that of solving their equivalent settings in Model RB. 
>From many previous studies \cite{gent,smith,prosser} 
we know that the instances generated at the phase transition in many
settings of Model B are very hard to solve for various kinds of CSP algorithms.
So there exist a lot of hard instances to solve in Model RB.

\section{Conclusions and Future Work}
A lot of experimental and theoretical studies indicate 
that a phase transition in solvability is a very
important feature of many decision problems in computer science. 
It is shown that these problems can be characterized 
by a control parameter in such a way that the space of problem instances is
divided into two regions: 
the under-constrained region where almost all problems have many
solutions, and the over-constrained region where almost all problems
have no solutions, with a sharp transition between them.
Another interesting feature associated with the phase transition is
that the peak in hardness of solving the problem instances occurs 
in the transition region. Since these instances generated in the 
transition region appear hardest to solve, they are commonly used
as a benchmark for algorithms for many NP-complete problems. 
But unfortunately, except for the Hamiltonian cycle problem 
(which is NP-complete), all the decision problems that have exact
results about the existence and the location of the phase transition 
are in P class \cite{parkes}, e.g.  random 2-SAT. 
These problems are not so interesting as the NP-complete problems
from a complexity theoretic point of view 
because they can be solved in polynomial time. 
For the Hamiltonian cycle problem, using an improved backtrack algorithm 
with sophisticated pruning techniques, Vandegriend and Culberson
\citeyear{vandegriend}
recently found that the problem instances in the
phase transition region are not hard to solve.

In this paper we proposed a new type of random CSP model, Model RB, 
which is a revision to the standard Model B, 
and the asymptotic analysis of this model has also been presented. The
results are quite surprising. We can not only prove the existence 
of phase transitions in this model but also know the location of transition
points exactly. It was further shown that there exist a lot of
hard instances in Model RB by relating its hardness to the standard Model B. 
Since there is still some lack of studies about the exact derivation of 
the phase transitions in NP-complete problems,
this paper may provide some new insight into this field. 
However, we did not discuss the scaling behaviour of Model RB and
some other related issues in this paper. In order to get a better
understanding of Model RB, we suggest that future work should include
determining either empirically or theoretically whether or not hard 
instances persist with reasonably high frequency
as the number of variables increases.
\footnote{Two anonymous referees suggest this point.}

\acks{This research was supported by National 973 Project of
China Grant No. G1999032701.
We would like to thank Ian P. Gent, Barbara M. Smith, 
Peter van Beek and the anonymous referees for their helpful 
comments and suggestions.}

\vskip 0.2in
\bibliography{exactph}
\bibliographystyle{theapa}
\end{document}